\documentclass{article}
\usepackage{spconf,amsmath,graphicx}
\usepackage{amsfonts}
\usepackage{algorithmic}
\usepackage{algorithm}
\usepackage{array}
\usepackage[caption=false,font=normalsize,labelfont=sf,textfont=sf]{subfig}
\usepackage{textcomp}
\usepackage{stfloats}
\usepackage{url}
\usepackage{verbatim}
\usepackage{hyperref}
\usepackage{caption}
\usepackage{multirow}
\usepackage{float}
\usepackage{svg}
\usepackage{graphicx,subfig}

\title{Multimodal-Enhanced Objectness Learner for Corner Case Detection in Autonomous Driving}
\name{Lixing Xiao$^{1,2}$, Ruixiao Shi$^{1,2}$, Xiaoyang Tang$^{1,2}$, Yi Zhou$^{1,2}$\sthanks{Corresponding author: Yi Zhou, yizhou.szcn@gmail.com}}
\address{$^{1}$School of Computer Science and Engineering, Southeast University, Nanjing, China \\
$^{2}$Key Laboratory of New Generation Artificial Intelligence Technology\\
and Its Interdisciplinary Applications (Southeast University), Ministry of Education}
\begin{document}
%
\maketitle
\begin{abstract}
  Previous works on object detection have achieved high accuracy in closed-set scenarios, but their performance in open-world scenarios is not satisfactory.
  One of the challenging open-world problems is corner case detection in autonomous driving.
  Existing detectors struggle with these cases, relying heavily on visual
  appearance and exhibiting poor generalization ability. In this paper, we propose a solution by
  reducing the discrepancy between known and unknown classes and introduce a
  multimodal-enhanced objectness notion learner. Leveraging both
  vision-centric and vision-language multiple modalities, our semi-supervised learning
  framework imparts objectness knowledge to the student model, enabling
  class-aware detection. Our approach, \textbf{M}ultimodal-\textbf{E}\textbf{n}hanced \textbf{O}bjectness \textbf{L}earner (MENOL)
  for Corner Case Detection, significantly improves recall for novel
  classes with lower training costs. By achieving a 76.6\% mAR-corner and 79.8\% mAR-agnostic on the CODA-val dataset with just 5100 labeled training images, MENOL outperforms the baseline ORE by 71.3\% and 60.6\%,
  respectively. The code will be available at \url{https://github.com/tryhiseyyysum/MENOL}.
\end{abstract}
\section{Introduction}
\label{sec:intro}
  Autonomous driving technologies pursue the goal of driving a vehicle without human intervention and guaranteeing the safety at the same time.
  Object detection is a fundamental task in autonomous driving, which aims to identify and localize objects in an image.
  The adoption of deep learning has accelerated progress in object detection research, many carefully designed deep neural networks have been proposed to improve the accuracy of object detection \cite{redmon2016you,erhan2014scalable}.
  The existing object detection methods have already achieved high accuracy in closed-set scenarios, where the object categories are pre-defined in advance.
  However, the performance of these methods in open-world scenarios is not satisfactory, where novel object categories and instances can be encountered.
  \begin{figure}[!t]
    \centering
    \includegraphics[width=1.0\columnwidth]{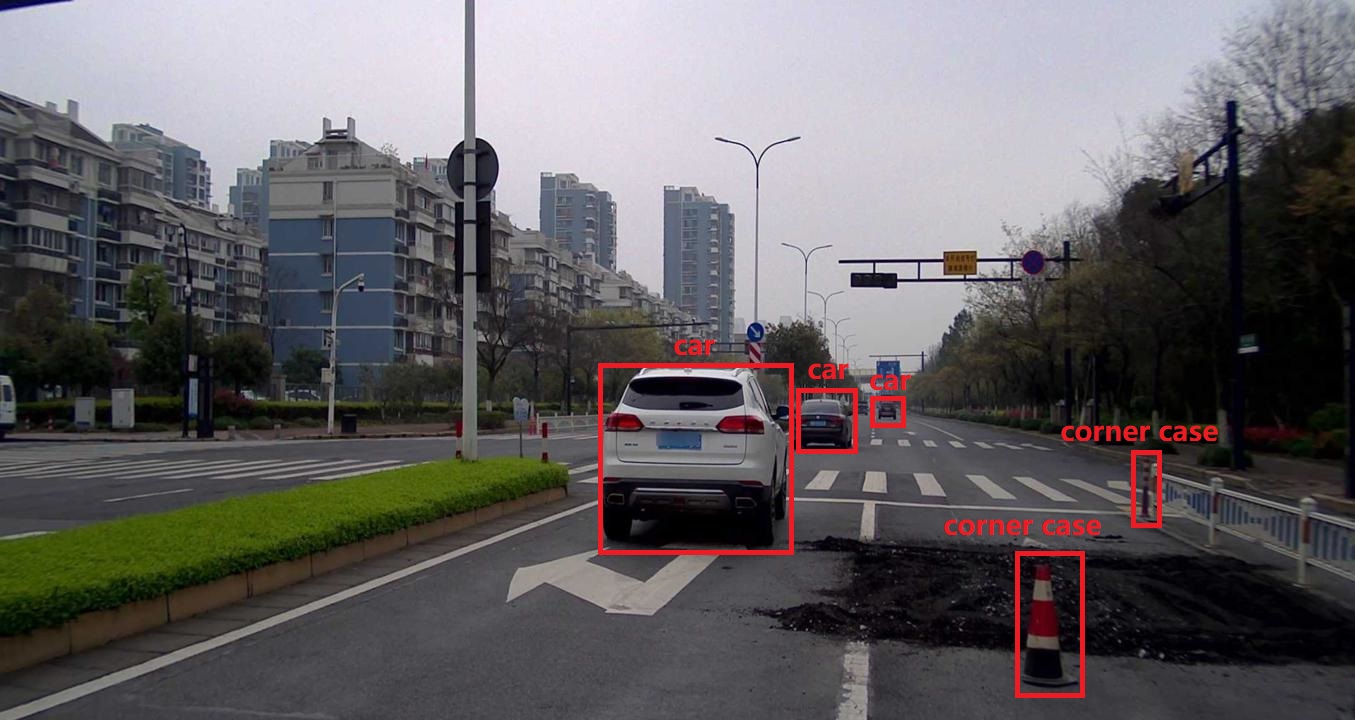}
    \caption{An example of corner case. The category of `traffic cone' is usually absent in most autonomous driving datasets.}
    \label{figure:corner}
  \end{figure}
  A main open-world object detection challenge in autonomous driving is called corner case detection.
  Corner case consists of two types:
  (i) instance of novel class (e.g., a runaway tire) and (ii) novel instance of common class (e.g., an overturned truck) \cite{li2022coda}.
  An example of corner case is shown as Fig.\ref{figure:corner}.
  However, most of the detectors are generally incapable of detecting novel objects that are not seen or rarely seen in the training process,
  resulting in low recall and high false negative rate.
  Object detection in autonomous driving may encounter corner case problem.
  One method to deal with this problem is called \textbf{O}pen \textbf{W}orld \textbf{O}bject \textbf{D}etection (OWOD). 
  The work of \cite{joseph2021towards} proposed this paradigm, which aims to detect a given set of known objects while simultaneously learn to identify unknown objects.
  ORE \cite{joseph2021towards} is the first open-world object detector. It utilizes a learnable energy-based unknown identifier to distinguish unknown classes from the known ones.
  OW-DETR \cite{gupta2022ow} extends ORE by using a transformer-based framework to explicitly address the OWOD challenges end-to-end.
  The works of OWOD have shown promising improvements over the closed-set object detectors,
  but they still have many limitations. Specifically, these fully-supervised methods have a strong bias towards known classes.
  As a result, they fail to detect the novel classes of corner case.
  Additionally, their generalization performance will be poor if the training dataset is not large enough.
  More recently, many studies based on multiple modalities have made
  significant progress. Diverse information contributes to the success of multiple modalities. It provides the model with various perspectives on understanding the world.
  The proposal of
  CLIP \cite{radford2021learning} indicates that image-text pairs contain boarder visual
  concepts than the pre-defined concepts.
  Visual representations can be learned on large amounts of image-text
  pairs. The introduction of the large-scale vision-language model CLIP has contributed to the development of \textbf{O}pen-\textbf{V}ocabulary \textbf{O}bject
  \textbf{D}etection (OVOD) and \textbf{Z}ero-\textbf{S}hot \textbf{O}bject \textbf{D}etection (ZSOD).
  Both of them
  focus on how to train an object detector on base classes and then
  generalize to novel classes during inference.
  The first work of OVOD \cite{zareian2021open} is based on captioning information extraction. 
  Most existing ZSOD methods like \cite{rahman2020improved} learn from the base classes and generate to novel classes
  by exploiting correlations between base and novel categories.     
  Although OVOD and ZSOD are effective for some open-world scenarios, we argue that they are not suitable for solving the corner case problem.
  The reason is that the semantic boundaries of categories are not clear enough and the differences between corner case categories are large.
  For example, the categories of `debris' and `misc' are not as clear as the categories of `car' or `pedestrian',
  making it hard to align the vision and language representation spaces for OVOD or ZSOD pipeline.\\
  \indent In contrast to ZSOD or OVOD, 
  we propose a \textbf{M}ultimodal-\textbf{E}\textbf{n}hanced \textbf{O}bjectness \textbf{L}earner (MENOL) for Corner Case Detection.
  MENOL mainly focuses on learning the notion of objectness and improving the recall of novel classes.
  Many studies have treated the corner case detection problem as an Out-Of-Distribution (OOD) \cite{nitsch2021out}
  or Anomaly Detection problem \cite{di2021pixel}.
  They design very complex rules to detect the corner case objects.
  However, the t-SNE visualization (Fig.\ref{figure:tsne}) shows that the distribution difference within the novel classes is large, making it difficult to design such general rules.
  We propose to narrow the gap between known and unknown classes by using geometry cues of depth and normal.
  To learn the notion of objectness, we design a multimodal-enhanced objectness notion learner,
  which is a vision-centric and vision-language multimodal design.
  Our MENOL is designed as a semi-supervised framework, the objectness notion learner acts as the teacher model to generate class-agnostic pseudo boxes for unlabeled data (the extracted depth and normal images).
  The pseudo-labeled depth and normal images are then merged with fully-annotated original RGB images and fed into a student model to train the final class-aware open-world object detector.
  Overall, our proposed MENOL successfully solves the problems mentioned above and significantly improves the recall of novel classes with relatively lower training cost.
  The idea of learning the notion of objectness and reducing the discrepancy between known and unknown classes is a general way to deal with the problem of corner case detection.\\
  \indent We highlight the contributions of this paper as follows:\\
  \indent \textbullet \indent We propose a novel multimodal-enhanced objectness notion learner and a semi-supervised based open-world object detector for corner case detection in autonomous driving called MENOL, which
  can effectively improve the recall of novel classes with relatively lower training cost. Our method provides a general way to reduce the discrepancy, which is simpler, lower-consumed, but also performs well in practice.\\
  \indent \textbullet \indent Vision-centric and vision-language multiple modalities are designed to provide additional knowledge and diverse information for the model to learn,
  enabling it to better understand the world and the notion of objectness.\\
  \indent \textbullet \indent We show that our MENOL outperforms many public baselines. Surpassing the ORE \cite{joseph2021towards} by 71.3\% and 60.6\% in terms of mAR-corner and mAR-agnostic respectively.

\section{Related work}
\label{sec:format}
\textbf{Closed-set Object Detection} in computer vision involves identifying and localizing pre-defined
objects in an image. Methods can be categorized as one-stage or two-stage.
One-stage methods like YOLO \cite{redmon2016you} directly perform localization and classification
in a single network. Two-stage methods, exemplified by Faster R-CNN \cite{ren2015faster}, have a proposal
extraction stage for candidate boxes generation and a subsequent stage for box
classification and refinement. While two-stage models offer higher accuracy,
they incur a greater computational cost due to the additional proposal generation stage.

\textbf{Open-world Object Detection} involves predicting category labels and bounding boxes for objects, including those with unknown categories that the model must learn.
Joseph et al.\cite{joseph2021towards} introduce ORE, which is the first open-world object detector using a region proposal network (RPN) to generate class-agnostic proposals.
It employs auto-labeling on pseudo-unknowns for training, distinguishing unknown classes using a learnable energy-based identifier.
Gupta et al.\cite{gupta2022ow} propose an end-to-end transformer-based framework to address open-world object detection challenges through attention-driven pseudo-labeling, novelty classification, and objectness scoring.
Huang et al.\cite{huang2022good} identify overfitting issues in RGB-based detectors and introduce GOOD, which is a novel framework leveraging geometry cues to enhance detection performance.

\textbf{Vision-Language Models (VLMs) and Open-Vocabulary Object Detection} involve detecting objects of unlearned categories by integrating multiple modalities. VLMs are large pre-trained models aligning image and text representations.
Contrastive Language-Image Pre-training (CLIP) \cite{radford2021learning} exhibits remarkable zero-shot ability in image classification.
Building on CLIP's success, researchers explore Open-Vocabulary Object Detection (OVOD) \cite{zareian2021open}, applying VLMs to detect objects of unseen categories. OVOD aligns individual visual embeddings with text embeddings, enabling strong correlations between them. For instance, Li et al.\cite{li2022grounded} introduce Grounded Language-Image Pre-training (GLIP), which is a large-scale vision-language model pre-trained at the region-word level.

\section{Method}
\label{sec:pagestyle}
\begin{figure*}[!t]
  \centering
  \includegraphics[width=\textwidth]{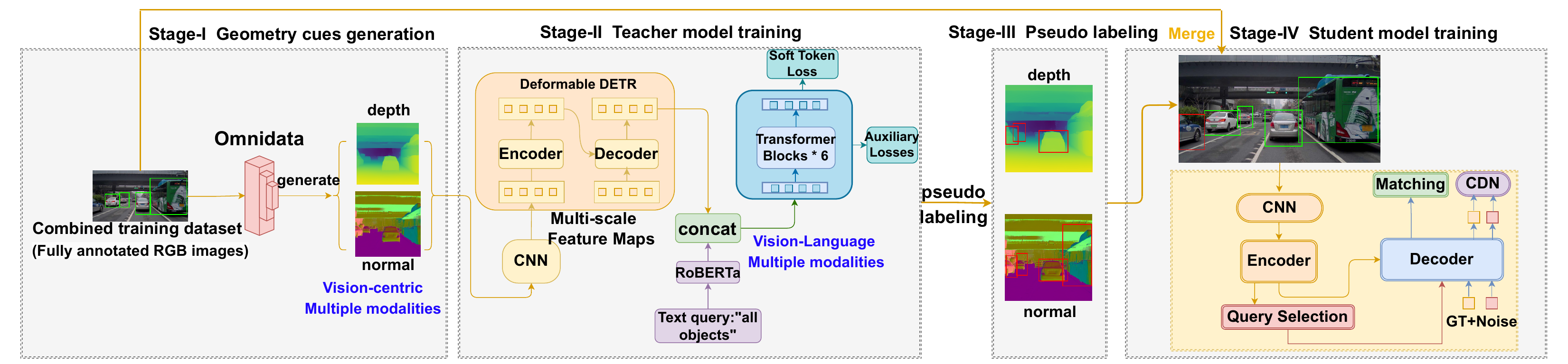}
  \caption{The overview of MENOL framework. It consists of 4 stages. (i) Stage I: the RGB images from training dataset are firstly pre-processed by the off-the-shelf Omnidata model to extract the geometry cues.
  (ii) Stage II: the generated geometry cues images are used to train the objectness notion learner.
  (iii) Stage III: the trained objectness notion learner is used as the teacher model to generate pseudo boxes for the depth and normal images from another autonomous driving dataset.
  (iv) Stage IV: the pseudo-labeled depth and normal images are merged with fully annotated original RGB images and then fed into the DINO-based student model to train the final class-aware open-world object detector.
  Only student model is used for inference.}
  \label{figure:model}
\end{figure*}
\indent Our MENOL provides a novel pipeline for detecting corner cases in autonomous driving by
leveraging multiple modalities and semi-supervised learning framework.
In order to reduce the discrepancy between known and unknown classes,
we utilize geometry cues of depth and normal to provide additional knowledge and diverse information for the model to learn,
which is a kind of vision-centric multiple modalities.
With the goal of learning the notion of objectness and improving the recall of novel classes,
we train an objectness notion learner to perform class-agnostic detection of all objects.
To infuse the knowledge of objectness notion into the student model,
we design a semi-supervised learning framework. The objectness notion learner acts as the teacher model to generate pseudo boxes for unlabeled data (the extracted depth and normal images).
Then, these pseudo-labeled depth and normal images are merged with fully-annotated original RGB images to train a student model, which is the final class-aware open-world object detector for the task of corner case detection.
The overview of our model is shown in Fig.\ref{figure:model}.

\subsection{Geometry Cues Extraction}
\indent Depth focuses on the relative spatial difference of objects, ignoring details on the object surfaces, while normal focuses on directional differences. Reducing the discrepancy between known and unknown objects in terms of geometry cues is beneficial for the model to learn the notion of objectness. The depth and normal images, along with the original RGB images, constitute vision-centric multiple modalities, preventing the model from overfitting training classes and relying solely on visual appearance cues when detecting objects.\\
\indent As shown in Fig.\ref{figure:model} Stage I, we use the Omnidata model \cite{eftekhar2021omnidata} to extract depth and normal images from the original RGB images in the training dataset.
Omnidata model is trained on the Omnidata Starter Dataset (OSD) \cite{eftekhar2021omnidata}, using cross-task consistency and 2D/3D data augmentations.
It can produce high-quality depth and normal images and the invariances behind these geometric cues are robust.

\subsection{Learning the Notion of Objectness}
Inspired by the previous work of GOOD \cite{huang2022good}, we also use geometry cues to enhance our open-world object detector's performance. GOOD trains a proposal network based on OLN \cite{kim2022learning},
using RGB images of known classes to generate proposals for depth and normal images.
However, GOOD has limitations.
Firstly, OLN just simply replaces the classifier head of Faster R-CNN \cite{ren2015faster} with localization quality estimators and 
uses relatively small scale of images to train it. As a result, the model still has a strong bias towards known classes and is not able to learn the notion of objectness.
It performs poorly on the task of corner case detection if the training dataset is not large enough.
Additionally, the proposal network of GOOD is trained on RGB images and required to generate proposals for depth and normal images,
the different characteristics of these two kinds of images limit the generalization performance of the proposal network.
Finally, GOOD can only perform class-agnostic object detection, which is not consistent with the practical application of autonomous driving.
Our approach addresses these limitations.\\
\indent To solve the problem of overfitting the known classes in the proposal network of GOOD,
we propose to use an objectness notion learner to perform class-agnostic detection of all objects.
Our objectness notion learner is a vision-language multimodal design. Deformable DETR architecture is used as the vision branch to extract the visual features,
and the RoBERTa is used as the language branch to encode the text.\\
\indent Since objects in an image are not always in the same scale, the scale variation is a great challenge for autonomous driving object detectors.
In our objectness notion learner, the image is first fed into a CNN backbone to extract multi-scale visual features.
As shown in Fig.\ref{figure:model} Stage II,
the extracted features are then fed into the Deformable DETR to obtain vision representation vectors.
The attention is calculated at multiple scales by the multi-scale deformable attention module to incorporate better contextual information.
To reduce the computational cost, the deformable attention samples a small set of keys around a reference (query) image location to achieve linear complexity with respect to the size of the image feature maps.
The pre-trained RoBERTa encodes the text query input and produces a corresponding sequence of hidden vectors.
Concatenating the flattened image features and the text embeddings would be likely to destroy the spatial structure of an image.
Hence, a better choice is to use a late multimodal fusion mechanism to fuse the image features and the text embeddings.
Specifically, the image features are first processed the Deformable DETR architecture to obtain the object query representations.
After concatenating with the text embeddings, they are fed into a Transformer architecture to fuse multimodal information.
The output head is applied after each transformer self-attention block and the soft token loss and auxiliary losses are calculated to optimize the parameters.
Soft token loss is designed for predicting the span of tokens from the original text that refers to each matched object,
instead of predicting a categorical class for each detected object.
The model is trained to predict a uniform distribution over all token positions that
correspond to the object for each predicted box that is matched to a ground truth box using the bi-partite matching.
The use of vision-language multiple modalities enables the model to learn from large amounts of image-text pairs and get a better understanding of objectness.
\subsection{Infusing Knowledge of Objectness Notion through Semi-supervised Learning Framework}
Previous works on semi-supervised object detection (SSOD) have achieved great success,
commonly employing pseudo labeling by a teacher model.
Recent study \cite{liu2022open} extends
SSOD to open-world object detection, outperforming fully supervised methods.

As the objectness notion learner focuses solely on `what is an object?',
it lacks class-aware detection capabilities. To leverage this knowledge for
class-aware detection, we introduce a semi-supervised learning framework.
As shown in Fig.\ref{figure:model} Stage III and IV,
the teacher model generates class-agnostic pseudo boxes for unlabeled data (the extracted depth and normal images), infusing the objectness notion knowledge into the student model.
Considering the diverse aspects reflected in these images,
the teacher model generates distinct pseudo boxes.
This design leverages vision-centric multiple modalities to provide diverse information for the student model.
Pseudo-labeled depth and normal images are merged with fully
annotated RGB images and used to train the student model for open-world class-aware object detection.

The student model adopts an end-to-end architecture based on DINO \cite{zhang2022dino} with a Swin Transformer backbone.
It employs contrastive denoising training, mixed query selection for anchor initialization,
and a look-forward twice scheme for box prediction.

\subsection{Optimization Objective}
The optimization objective of our student model includes classification loss and box regression loss.
The classification loss is general Focal Loss.
The box regression loss is calculated as a combination of the generalized intersection over union (GIoU) loss and the L1 loss:
\begin{equation}
  \begin{aligned}
    \mathcal{L}_{\text{box}}(b_i,\hat{b_i}) = \lambda_{\text{1}}\mathcal{L}_{\text{\text{GIoU}}}(b_i,\hat{b_i}) + \lambda_{\text{2}}||b_i - \hat{b_i}||_1 ,\\
  \end{aligned}
  \label{equation:loss}
\end{equation}
where $b_i$ and $\hat{b_i}$ are the vectors that define ground truth and predicted boxes, $\lambda_{\text{1}}$ and $\lambda_{\text{2}}$ are hyperparameters.\\
\indent The GIoU loss is calculated as:
\begin{equation}
  \begin{aligned}
    \mathcal{L}_{\text{\text{GIoU}}}(b_i,\hat{b_i}) = 1 - (\frac{|b_i \cap \hat{b_i}|}{|b_i \cup \hat{b_i}|} - \frac{|B(b_i,\hat{b_i}) \backslash b_i \cup \hat{b_i}|}{|B(b_i,\hat{b_i})|}). \\
  \end{aligned}
  \label{equation:giou}
\end{equation}
where $|\cdot|$ means `area' and $B(b_i,\hat{b_i})$ is the largest bounding box that encloses both $b_i$ and $\hat{b_i}$.

\section{Experiments}
\label{sec:typestyle}
\subsection{Datasets and Evaluation Metrics}
\begin{figure}[t]
  \centering
  \includegraphics[width=1.0\columnwidth]{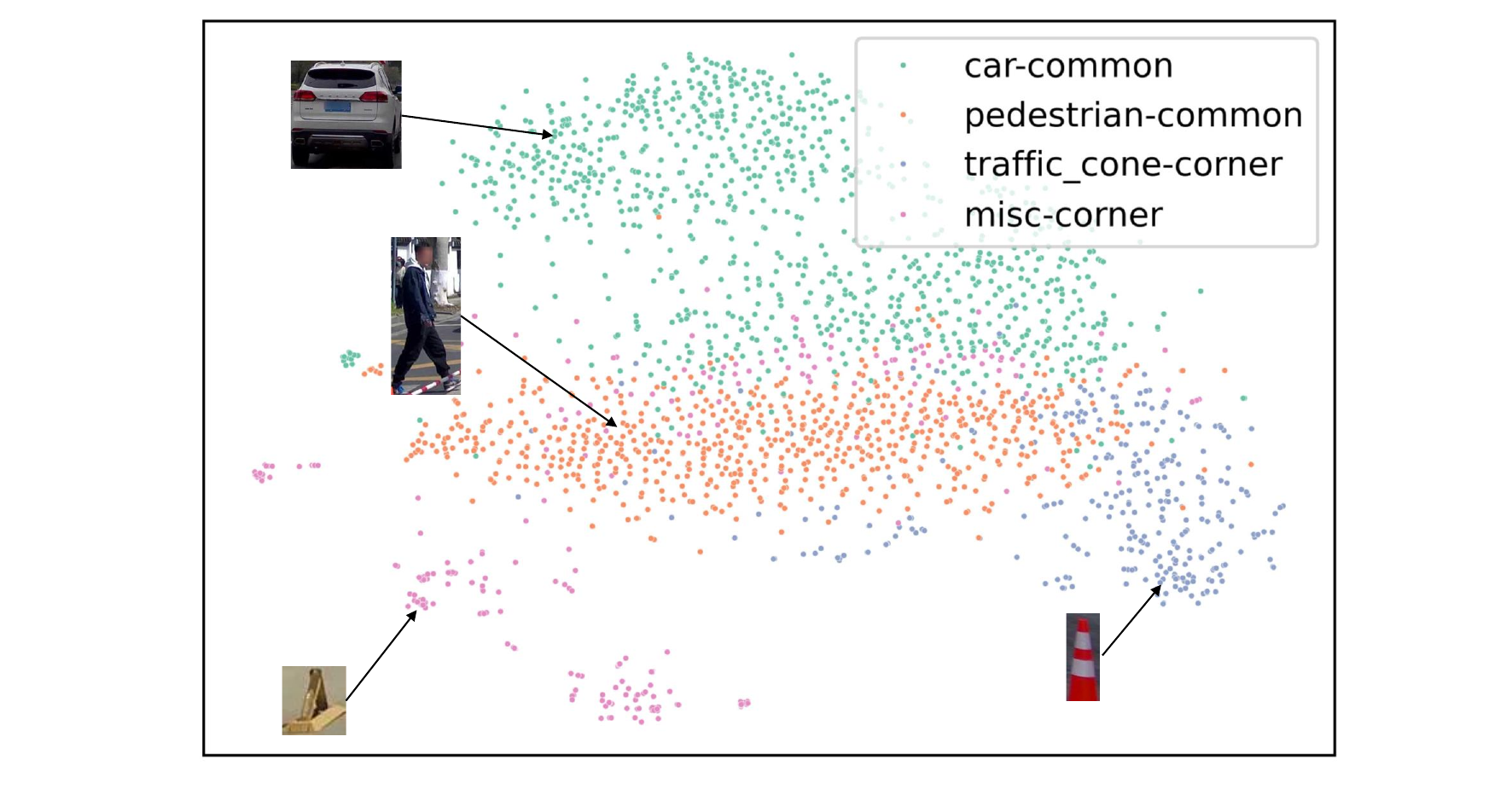}
  \caption{The t-SNE visualization of CODA \cite{li2022coda}. Two common classes and two corner classes are chosen for visualization.}
  \label{figure:tsne}
\end{figure}

We conduct training on the combination of CODA \cite{li2022coda} and SODA10M \cite{han2021soda10m} dataset.
CODA is a novel dataset of object-level corner cases in real world scenes,
consisting of approximately 10k carefully selected road driving scenes with
high-quality bounding box annotations for 29 representative object categories.
SODA10M is a 2D object detection dataset with 6 representative categories.
We sample 100 images from CODA dataset and combine them with the SODA10M-labeled training dataset as our training dataset.
Then we sample another 488 images from CODA dataset and combine them with the SODA10M-labeled validation dataset as our validation dataset.
The model is tested on the CODA-val \cite{li2022coda} dataset (not overlap with the aforementioned datasets) and BDD100K-val \cite{yu2020bdd100k} dataset to evaluate its corner case detection and common object detection performance respectively.
It simulates the real-world scenario where novel categories will be encountered.
We use t-SNE \cite{vandermaaten08a} to visualize the distribution of the images in the training dataset.
As shown in Fig.\ref{figure:tsne},
the pre-trained ResNet-34 is used for feature extraction and the parameters of its fourth layer are extracted for t-SNE dimension reduction. The result shows that the corner cases vary considerably both within and between classes.\\
\indent The categories are divided into common and novel classes.
Seven selected categories are considered as common classes (i.e., pedestrian, cyclist, car, truck, tram, tricycle, bus), and the rest 22 categories are novel classes.
We detect the 7 common classes in a class-aware manner and the novel classes in a class-agnostic manner,
which means that all novel classes are mapped to the same category ID during inference.
The following 4 metrics are used to evaluate the performance of our model:
(i) \textbf{AR-agnostic-corner}: mAR over corner-case objects of all categories in a class-agnostic manner;
(ii) \textbf{AR-agnostic}: mAR over objects of all categories in a class-agnostic manner;
(iii) \textbf{AP-agnostic}: mAP over objects of all categories in a class-agnostic manner;
(iv) \textbf{AP-common}: mAP over objects of common categories.
\subsection{Baselines}
We choose Faster R-CNN \cite{ren2015faster}, ORE \cite{joseph2021towards}, OW-DETR \cite{gupta2022ow} and
GOOD \cite{huang2022good} as the baseline method for comparison.
The original GOOD model is a class-agnostic object detector, which means the detection results are just foreground or background.
For a fair comparison, we replace the OLN-based RPNHead, ROIHead and classification head in GOOD with the standard versions of Faster R-CNN respectively,
so as to make it perform class-aware object detection.

\subsection{Implementation Details}
We extract geometry cues (depth and normal) from the original RGB images in the training set of CODA and SODA10M using Omnidata model \cite{eftekhar2021omnidata}.
They are unlabeled but we can use the annotations of their corresponding RGB images as their labels.
Then, BLIP \cite{li2022blip} is used to generate captions for the images from CODA training dataset.
Depth and normal images share the same captions as the original RGB images.\\
\indent The CNN backbone of the objectness notion learner is initialized with the weights of ResNet-101 pre-trained on ImageNet-1K.
The objectness notion learner is pre-trained using approximately 1.3M aligned image-text pairs from MS COCO, Flickr30k and Visual Genome (VG).
These datasets contain various objects from different scenarios. After pre-training, the model has learned the notion of objectness to some extent.
A combination of extracted depth and normal images and RGB images in CODA dataset,
together with their corresponding captions are fed into the objectness notion learner to fine-tune the model.
In the fine-tuning stage, the parameters of the CNN backbone, the Deformable DETR and the language backbone are frozen while those of the late fusion Transformer are updated.
Since we are not concerned about the specific categories of objects, and just focus on `what is an object?',
we use `all objects' as the text query to generate class-agnostic predictions in the inference stage.\\
\indent The trained objectness notion learner acts as the teacher model to generate pseudo boxes for the extracted depth and normal images in our training dataset (i.e. SODA10M).
The pseudo-labeled depth and normal images are merged with fully annotated original RGB images and fed into the student model.
The student model is a closed-set object detector based on DINO \cite{zhang2022dino}.
It is pre-trained on Objects365 dataset (about 1.7M annotated images) with a Swin Transformer-large backbone.
The hyperparameters of the loss function \ref{equation:loss} are set as $\lambda_{\text{1}}=2.0$ and $\lambda_{\text{2}}=5.0$.
We use AdamW optimizer with initial learning rate of 0.0001 and weight decay of 0.0001 to train the DINO-based student model.
The training process accomplishes in 35 epochs using 2 NVIDIA Geforce RTX 3090 GPUs with batch size of 2.

\vspace{-0.2em}

\subsection{Results}
\begin{figure}[t]
  \centering
  \begin{minipage}{0.4\textwidth}
    \includegraphics[width=1.0\columnwidth]{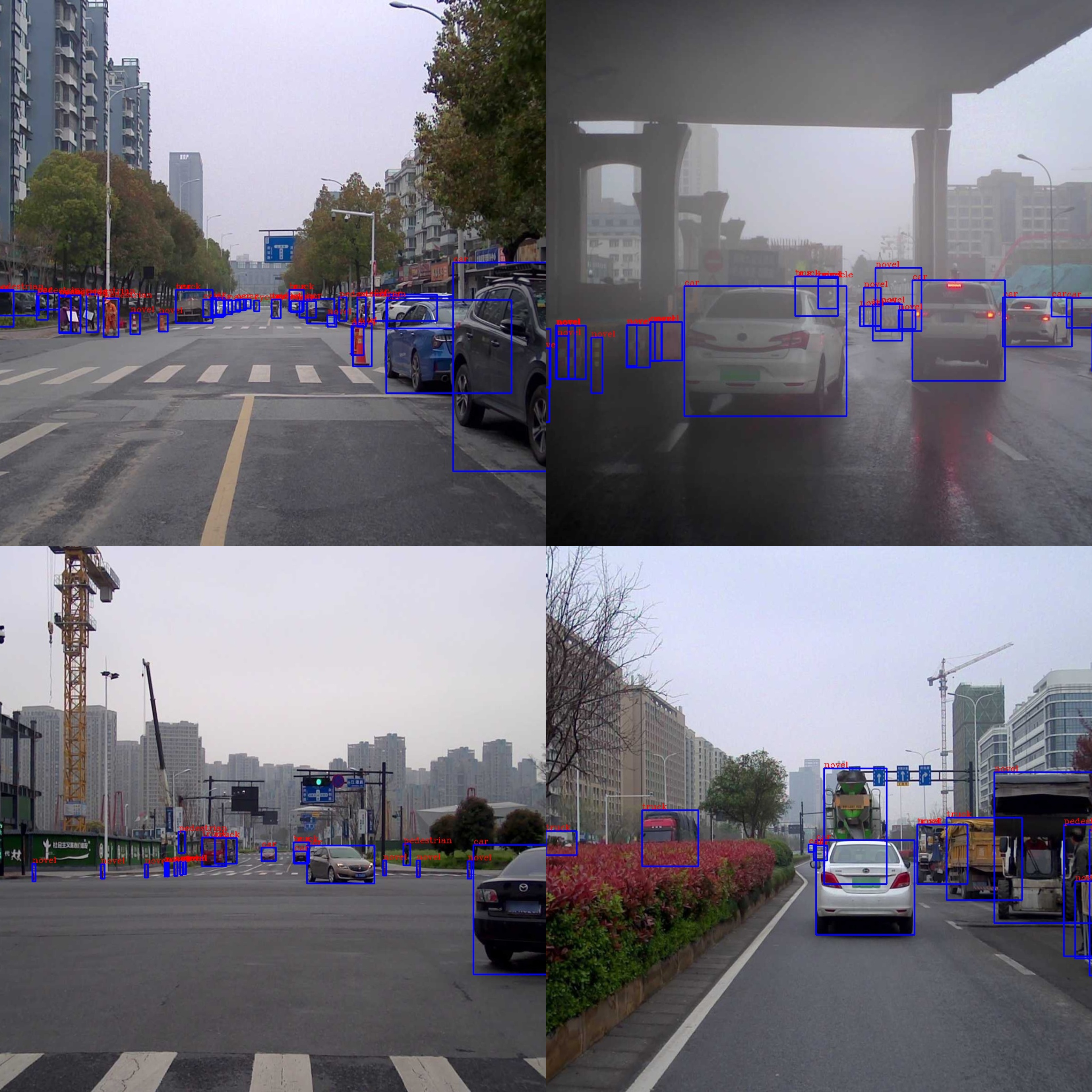}
  \end{minipage}
  \caption{Detection result of our MENOL on CODA \cite{li2022coda} dataset.}
  \label{figure:combine}
\end{figure}

The performance of our MENOL and the baseline methods on the CODA-val dataset is shown in Table \ref{table:results}.
Our MENOL achieves 0.766 mAR-corner, 0.798 mAR-agnostic, 0.742 mAP-agnostic and 0.711 mAP-common,
which outperforms the baseline models by a large margin.
This indicates that our MENOL has a better corner case detection ability.
The performance of our MENOL and the baseline methods on the BDD100K-val dataset are shown in Table \ref{table:results2}.
Our MENOL achieves 0.882 Recall and 0.786 mAP50, outperforming the baseline models.
This indicates that our MENOL still has a better common object detection ability.
The detection result of our MENOL on the CODA dataset is shown in Fig.\ref{figure:combine}.

\begin{table}[H]
  \centering
  \begin{minipage}{0.5\textwidth}
    \centering
    \resizebox{\textwidth}{!}{
    \begin{tabular}[H]{lcccc}
      \hline
      Method & AR-agnostic-corner & AR-agnostic & AP-agnostic & AP-common \\
      \hline
      Faster R-CNN \cite{ren2015faster} & 0.040 & 0.233 & 0.174 & 0.182 \\
      ORE \cite{joseph2021towards} & 0.053 & 0.192 & 0.371 & 0.284 \\
      OW-DETR \cite{gupta2022ow} & 0.432 & 0.464 & 0.363 & 0.242 \\
      GOOD \cite{huang2022good} & 0.585 & 0.628 & 0.564 & 0.340 \\
      \textbf{MENOL}(\textbf{Ours}) & \textbf{0.766} & \textbf{0.798} & \textbf{0.742} & \textbf{0.711} \\
      \hline
    \end{tabular}
    }
    \caption{Comparison of our MENOL with the baseline models on the CODA-val \cite{li2022coda} dataset.}
    \label{table:results}
  \end{minipage}
\end{table}

\subsection{Ablation Studies}
\begin{table}[t]
  \centering
  \begin{minipage}{0.5\textwidth}
    \centering
    \resizebox{0.65\textwidth}{!}{
    \begin{tabular}[t]{lcc}
      \hline
      Method & Recall & mAP50 \\
      \hline
      Faster R-CNN \cite{ren2015faster} & 0.772 & 0.556 \\
      MultiNet \cite{wu2022yolop} & 0.813 & 0.602 \\
      YOLOv5s \cite{wu2022yolop} & 0.868 & 0.772 \\
      \textbf{MENOL}(\textbf{Ours}) & \textbf{0.882} & \textbf{0.786} \\
      \hline
    \end{tabular}
    }
    \caption{Comparison of our MENOL with the baseline methods on the BDD100K-val \cite{yu2020bdd100k} dataset.}
    \label{table:results2}
  \end{minipage}
\end{table}

We conduct 3 ablation studies to verify the effectiveness of geometry cues, objectness notion learner and DINO-based student model.
Table \ref{table:ablation} shows the performance results of using RGB-only, Geometry cues-only and RGB+Geometry cues in the objectness notion learner respectively.
The RGB-only objectness notion learner outperforms the Geometry cues-only objectness notion learner,
indicating that geometry cues can not simply replace RGB images.
The RGB+Geometry cues objectness notion learner achieves the best performance,
which shows that the vision-centric multiple modalities is beneficial for the model to learn the notion of objectness.
The geometry cues help discover novel-looking objects that
RGB-based detectors cannot detect, and the RGB images have stronger visual semantic representations.
The combination of RGB and geometry cues can make up for the shortcomings of each other and achieve better performance.

\begin{figure}[t]
  \centering
  \includegraphics[width=0.8\columnwidth]{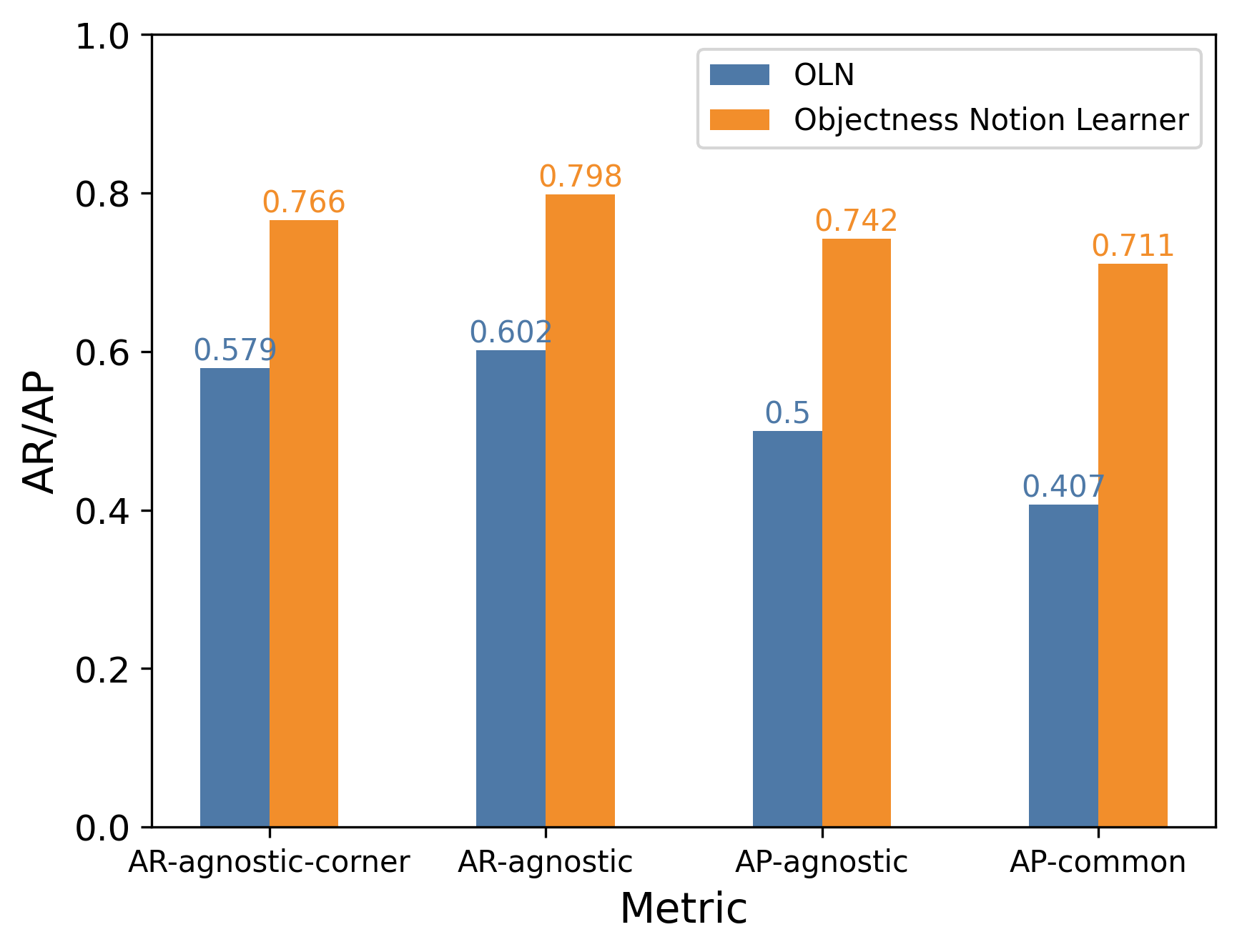}
  \caption{Comparison of our objectness notion learner with the proposal network of OLN \cite{kim2022learning} on the CODA-val \cite{li2022coda} dataset.}
  \label{figure:proposal}
\end{figure}

We further study the effectiveness of our objectness notion learner by comparing it with the proposal network of OLN \cite{kim2022learning}.
Fig.\ref{figure:proposal} shows the performance results.
The OLN-based proposal network performs poorly while our objectness notion learner achieves better performance,
indicating that the notion of objectness is more valuable than simple proposals.

\begin{figure}[t]
  \centering
  \includegraphics[width=0.8\columnwidth]{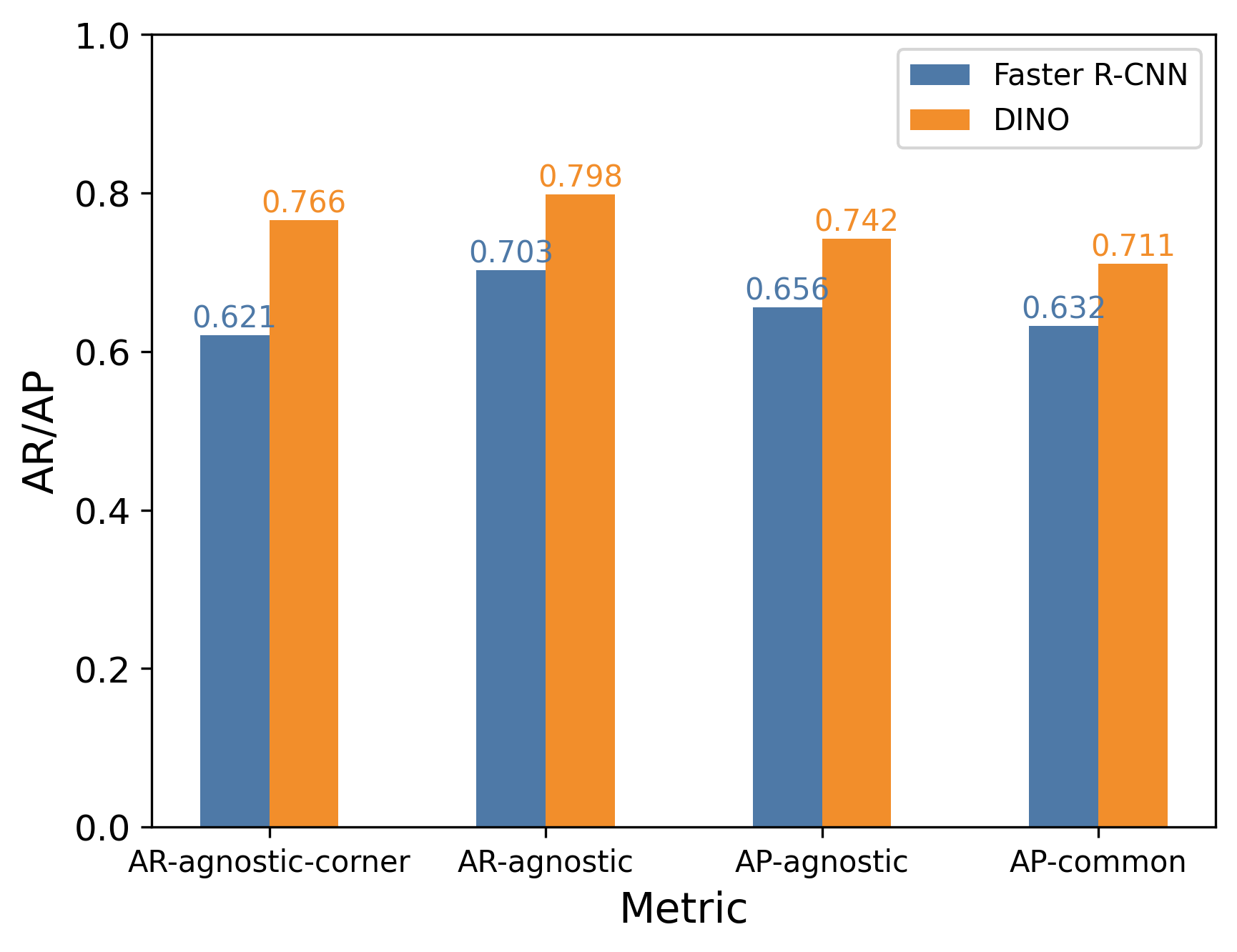}
  \caption{Comparison of our DINO-based \cite{zhang2022dino} student model with the Faster R-CNN-based \cite{ren2015faster} student model on the CODA-val \cite{li2022coda} dataset.}
  \label{figure:student}
\end{figure}

Finally, we study the effectiveness of our DINO-based \cite{zhang2022dino} student model by comparing it with Faster R-CNN-based \cite{ren2015faster} student model.
As shown in Fig.\ref{figure:student},
DINO-based student model outperforms Faster R-CNN-based student model,
which indicates that DINO-based student model is better.

\begin{table}[H]
  \centering
  \begin{minipage}{0.5\textwidth}
    \centering
    \resizebox{\textwidth}{!}{
    \begin{tabular}[H]{cc|cccc}
      \hline
      \multicolumn{2}{c|}{Image modalities} & \multicolumn{4}{c}{Metrics} \\
      \hline
      RGB & Geometry cues & AR-agnostic-corner & AR-agnostic & AP-agnostic & AP-common \\
      \hline
      \checkmark & & 0.557 & 0.571 & 0.446 & 0.333 \\
       & \checkmark & 0.352 & 0.387 & | & | \\
      \checkmark & \checkmark & \textbf{0.766} & \textbf{0.798} & \textbf{0.742} & \textbf{0.711} \\
      \hline
    \end{tabular}
    }
    \caption{Ablation studies of using RGB-only, Geometry cues-only, and RGB+Geometry cues in the objectness notion learner.}
    \label{table:ablation}
  \end{minipage}
\end{table}

\section{Conclusion}
\label{sec:majhead}
In conclusion, our MENOL leverages vision-centric and vision-language multimodal learning,
incorporating geometry cues to address corner case detection challenges in autonomous driving.
By effectively reducing the discrepancy between known and unknown classes,
MENOL demonstrates significant improvements in the recall of novel classes.
While our approach has shown promising results,
future work could explore richer modalities and more generalized objectness notion learning models to enhance the performance of road detection in autonomous driving.

\vfill\pagebreak

\bibliographystyle{IEEEbib}
\bibliography{refs}

\end{document}